\documentclass[final]{cvpr}

\usepackage{times}
\usepackage{epsfig}
\usepackage{graphicx}
\usepackage{amsmath}
\usepackage{amssymb}
\usepackage[acronym]{glossaries}
\usepackage{acronym}
\usepackage{lipsum}  
\usepackage{caption}
\usepackage{tabularx}
\usepackage{adjustbox}
\usepackage{booktabs}
\usepackage{graphicx}
\usepackage{caption}
\usepackage{subcaption}
\usepackage{blindtext}
\usepackage{epigraph}

\usepackage{xcolor}

\usepackage[ruled,vlined]{algorithm2e}

\SetCommentSty{mycommfont}

\usepackage{graphicx}

\acrodef{hmr}[HMR]{Human Mesh Recovery}
\acrodef{slp}[SLP]{Simultaneously-collected multimodal Lying Pose}
\acrodef{smpl}[SMPL]{Skinned Multi-Person Linear}
\acrodef{mpjpe}[MPJPE]{Mean Per Joint Position Error}
\acrodef{soa}[SOTA]{state-of-the-art}
\acrodef{spin}[SPIN]{SMPL Optimization IN the loop}

\usepackage[pagebackref=true,breaklinks=true,colorlinks,bookmarks=false]{hyperref}

\def\cvprPaperID{10263} 
\def\confYear{CVPR 2021}

\SetKwInput{KwInput}{Input}                
\SetKwInput{KwOutput}{Output}              

\begin{document}

\title{
Multimodal In-bed Pose and Shape Estimation under the Blankets
}

\author{Yu Yin, Joseph P. Robinson, Yun Fu\\
Department of Electrical and Computer Engineering\\
Northeastern University, Boston, MA\\
{\tt\small \{yin.yu1, robinson.jo\}@northeastern.edu, yunfu@ece.neu.edu}
}

\maketitle

\begin{abstract}
Humans spend vast hours in bed-- about one-third of the lifetime on average. Besides, a human at rest is vital in many healthcare applications. Typically, humans are covered by a blanket when resting, for which we propose a multimodal approach to uncover the subjects so their bodies at rest can be viewed without the occlusion of the blankets above.
We propose a pyramid scheme to effectively fuse the different modalities in a way that best leverages the knowledge captured by the multimodal sensors. Specifically, the two most informative modalities (\ie depth and infrared images) are first fused to generate good initial pose and shape estimation. Then pressure map and RGB images are further fused one by one to refine the result by providing occlusion-invariant information for the covered part, and accurate shape information for the uncovered part, respectively. 
However, even with multimodal data, the task of detecting human bodies at rest is still very challenging due to the extreme occlusion of bodies.
To further reduce the negative effects of the occlusion from blankets, we employ an attention-based reconstruction module to generate uncovered modalities, which are further fused to update current estimation via a cyclic fashion.
Extensive experiments validate the superiority of the proposed model over others.

\end{abstract}


\begin{figure}[t]
\begin{center}
   \includegraphics[width=\linewidth]{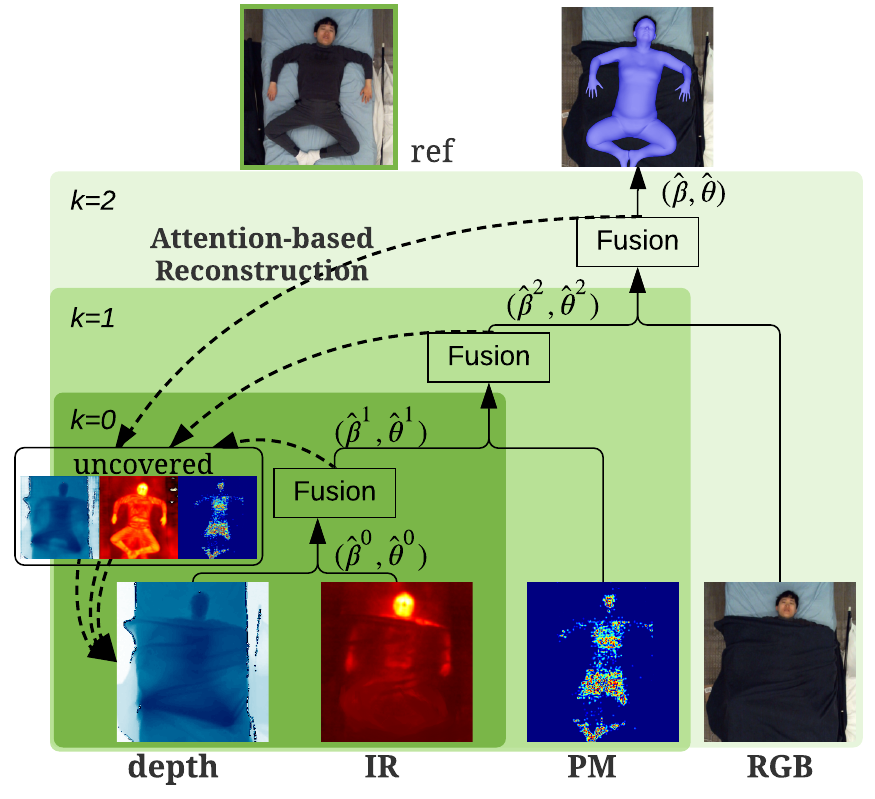}
\end{center}
   \caption{\textbf{Pyramid fusion scheme.} The $K+1$ levels fuse the multiple modalities in a sequence ordered by the most informative modality to the least (Table~\ref{tbl:result_ablation_single_mod}).}
\label{fig:pyramid}
\end{figure}

\begin{figure*}[t]
\centering
\begin{subfigure}[t]{0.23\textwidth}
  \includegraphics[width=\linewidth]{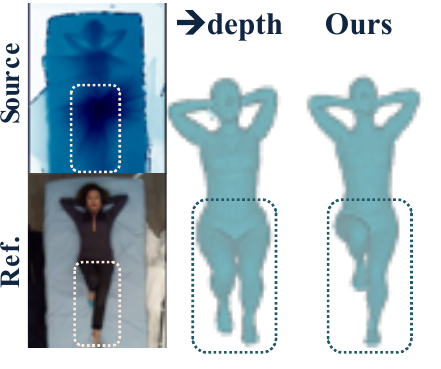}
\caption{}
\label{fig:edge_cases:depth}
\end{subfigure}
\begin{subfigure}[t]{0.23\textwidth}
  \includegraphics[width=\linewidth]{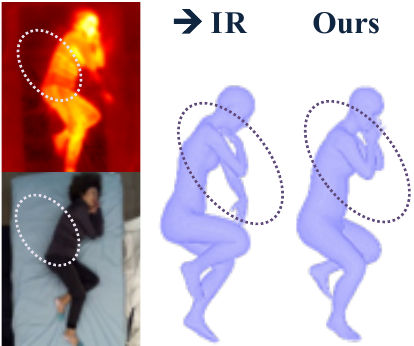}
\caption{}
\label{fig:edge_cases:IR}
\end{subfigure}
\begin{subfigure}[t]{0.22\textwidth}
  \includegraphics[width=\linewidth]{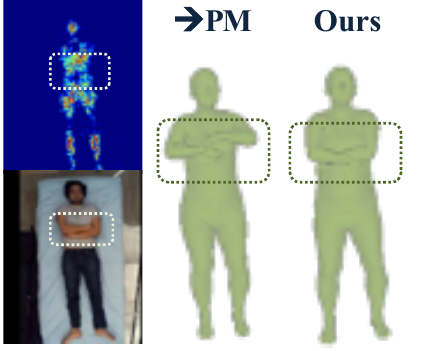}
\caption{}
\label{fig:edge_cases:PM}
\end{subfigure}
\begin{subfigure}[t]{0.22\textwidth}
  \includegraphics[width=\linewidth]{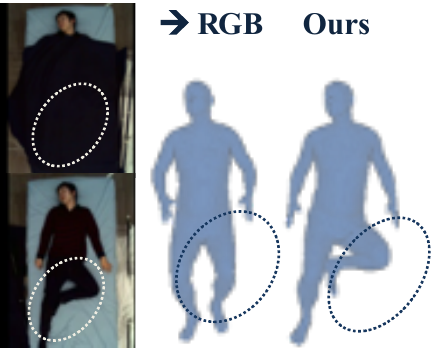}
\caption{}
\label{fig:edge_cases:RGB}
\end{subfigure}
  \caption{\textbf{Edge cases for the different modalities.} Shown are cases where the occlusion negatively affects the depth image (\emph{a}); heat residue derived from the previous pose causes an alias in the current IR frame (\emph{b}); body part is off the surface, leading to missing information in PM (\emph{a}); (d) RGB suffers from blanket occlusion (\emph{d}).}
\label{fig:edge_cases}
\end{figure*}


\section{Introduction}



Times that our bodies are at rest are critical-- sleep is imperative for human physical and mental health. 
Nowadays, researchers from the computer vision community have stepped up to research systems capable of acquiring an automatic understanding of bodies at rest, as seen fit in healthcare setting monitoring a patient long-term~\cite{chen2018patient}, managing cases of bedsores~\cite{pouyan2016automatic}, and even furthering our understanding via controlled \emph{sleep} studies~\cite{liu2019bed}. Many existing work in human \emph{in-bed} pose understanding mostly inherit technology proposed for 2D or 3D human pose estimation~\cite{casas2019patient, clever20183d, liu2019bed}. However, the aforementioned works lack critical knowledge for the shape of the body. We propose to estimate the shape of the body, as well as the pose, when at rest-- more similar to ~\cite{clever2020bodies}; however, the proposed framework clearly outdoes its predecessors with respect to the metrics and the overall vastness of information predicted (\ie access to the estimated 3D human mesh is a byproduct).


The task of detecting human bodies at rest (\ie \emph{in-bed}) raises additional challenges when compared to the traditional pose estimation problem: lights tend to be \emph{off} and bodies are typically covered in blankets, often entirely. From this, solutions for \emph{in-bed} pose estimation are limited in potential when in cases that use a single modality-- whether pressure map (PM)~\cite{casas2019patient, clever2020bodies, heydarzadeh2016bed}, infrared (IR)~\cite{liu2019seeing}, depth~\cite{achilles2016patient}, or RGB~\cite{liu2019bed}, the different modalities each come with different benefits and drawbacks (Figure~\ref{fig:edge_cases}). For instance, pressure maps are vulnerable to missing information: for example, a body part is off the surface for which the sensors measure the pressure (Figure~\ref{fig:edge_cases:PM}). In the end, the pressure signal is inaccurate with its limited reception field (\ie only spans the physical touch on the top of the mattress). Most research in general human pose estimation use RGB images~\cite{kanazawa2018end, kolotouros2019learning}. However, RGB has known flaws in the clinical setting: firstly, most or all of the human body is covered with a blanket (Figure~\ref{fig:edge_cases:RGB}), while in a room that often has no lighting and window shades to block out the sun; secondly, concerns in privacy are raised by many when being monitored by a visual signal that most all humans can interpret, which is especially the case for RGB videos. On the other hand, depth and IR images are more informative and are insensitive to change in lighting, or coverage. Still, they has their own drawbacks: some cases of occlusion negatively effect the depth images (Figure~\ref{fig:edge_cases:depth}), while IR can have errors where the heat residue derived from the previous pose as an alias for the current frame~\cite{liu20120simultaneously} (Figure~\ref{fig:edge_cases:IR}). 

In this paper, multimodal data is used to recover as much information as possible from bodies laying \emph{in-bed} and in the dark. To the best of our knowledge, we are the first to estimate 3D body poses and shapes in cases of such extreme occlusion. For this, we use novel fusion techniques to merge the multimodal data in such a way that maximizes the complimentary information encapsulated by the different modalities. In other words, we do not just feed the different modalities to the model simultaneously, but we have designed a pyramid structure that fuses a pair of modalities at a time, which allows the learned knowledge to be refined at each level that another modality is added. Specifically, we fuse the two most informative modalities (\ie depth and IR) at the bottom-most level of the pyramid, which to no surprise are the data types that best handle occlusion and are invariant to the lighting condition. Then, PM is passed in at the next level to provide its occlusion-invariant view of information (\ie contact area of body and mattress top). This allows the model to correct the estimation of the covered part well. Ultimately, the RGB is fused at the final, top-most level.
This level provides additional details and an accurate shape for the uncovered part.
Practically speaking, we do not always have access to all four modalities (\ie RGB, IR, depth, and PM). With the pyramid structure, we are able to still make good \emph{in-bed} pose estimations, even if only fewer modalities are available (\eg IR and depth) during testing. 


Even with multimodal data, the task of detecting \emph{in-bed} pose is still challenging due to the extreme occlusion of bodies. To reduce the noise information introduced by the blanket, we propose a coarse-to-fine estimation process using a novel attention-based \emph{reconstruction module}.
Specifically, a coarse 3D mesh estimation is first generated from input images. Then an attention-based reconstruction module is employed on the coarse mesh to generate the uncovered modalities, which are further fused to update the current estimation in a cyclic fashion.

In summary, we make the following contributions:
\begin{enumerate} 
    \item The first work to infer \emph{in-bed} poses and shapes from multiple modalities. A pyramid scheme is proposed to fuse the different modalities such to best leverage the knowledge captured by the different modalities. 
    \item We propose an attention-based \emph{reconstruction module} to further reduce the negative impact of the occlusion from blankets, and hence, improve the estimation of \emph{in-bed} poses, especially for occluded bodies.  
    
    \item We claim state-of-the-art and provide extensive analysis to study individual contributions of each modality for \emph{in-bed} pose and shape estimation. 
\end{enumerate}

\begin{figure*}[t!]
\begin{center}
\includegraphics[width=\textwidth]{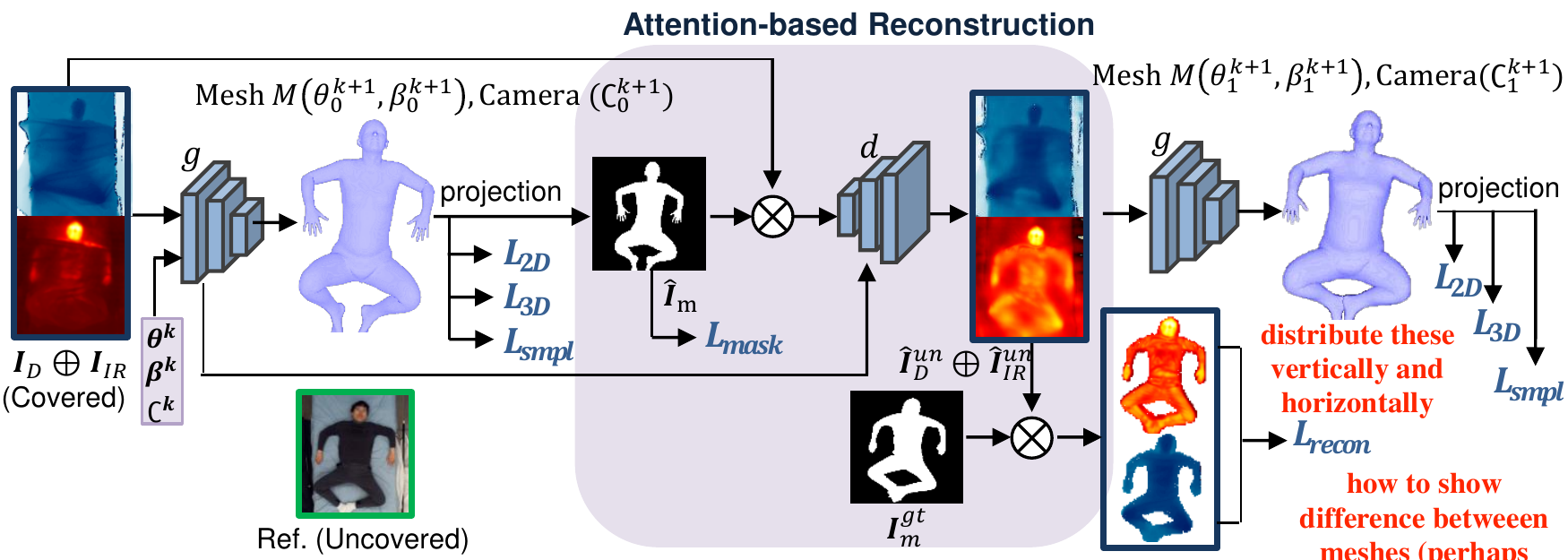}
\end{center}
   \caption{\textbf{Proposed framework depicting a single fusion level (\eg \eg $K=0$).}, The depth and IR are fused to predict \emph{in-bed} pose and shape parameters using a coarse-to-fine estimation process.}
\label{fig:proposedframework}
\end{figure*}

\section{Related Work}\label{sec:relatedworks}
The goal of human pose estimation is to localize human joints (or keypoints), \ie rigid bodies consist joints and rigid parts detectable by either or, and a body with clear pronunciation is a body with strong contortion. The task is then to localize the specific pose provided the possibility of all articulated poses. The number of keypoints varies across datasets (\eg LSP~\cite{andriluka20142d} has 14, MPII~\cite{sapp2013modec} has 16). The problem then classifies as one of two targets, 2D joints or 3D meshes. For 2D pose estimation, the target is ($x$, $y$) coordinates per joint in pixel space of an RGB image; For 3D pose estimation, the target is ($x$,$y$,$z$) coordinates in metric space of an RGB image. Depth information could too be present, in which case ($x$,$y$,$d$) or ($x$,$y$,$z$, $d$) as the target, which is multimodal as is the case here (\ie RGB, depth, IR, and pressure maps are used in this work).

We next cover the work done in pose estimation, and then the work specific to \emph{in-bed} poses. Finally, we review the literature on multimodal pose estimation.

\subsection{Human pose and shape estimation}
Pose estimation, in essence, is a classic problem. Researchers have proposed solutions for 2D~\cite{dang2019deep, munea2020progress}, 3D~\cite{sarafianos20163d,yang20183d}, and both~\cite{perez2014survey}. Earlier on, many focused on detecting human joints to then model and predict motion in a motion capture setting~\cite{moeslund2001survey, moeslund2006survey}. More recently, with the release of the Kinect sensor enabling researchers to acquire RGB-D in a controlled laboratory setting, the depth modality was leveraged to make enhanced, multimodal decisions~\cite{lun2015survey, ye2013survey}. Although not the focus of this work, pose estimation using human data is also done on targets such as human heads~\cite{murphy2008head, robinson2019laplace} and hands~\cite{li2019survey}, with, of course, commonalities in the proposed solutions.

Optimization-based approaches such as SMPlify~\cite{bogo2016keep} fit \ac{smpl} model~\cite{10.1145/2816795.2818013} to 2D keypoints.
Later, an end-to-end deep learning based regression approach (\ie, \ac{hmr}) is proposed to reconstruct entire 3D meshes of a human body given an RGB image~\cite{kanazawa2018end}.
Recently, Kolotouros~\etal proposed to combine the aforementioned, \ie, optimization-based learning and regressive shape fitting~\cite{kolotouros2019learning}.

\subsection{Human pose at rest}
Predicting poses of subjects in a bed and under the covers is relatively a new problem~\cite{liu20120simultaneously, liu2019seeing}. The RGB-D modalities were used to recognize a set of twenty humans when under the cover per a fuzzy logic machine learning algorithm~\cite{ren2020human}. Clever~\etal predicted pressure mappings of a subject in bed following a clever scheme based on synthetic training data~\cite{clever2020bodies}. Specifically, the authors used a physics simulation to synthesize data using a ragdoll model placed on top of flat fabric pressure sensors-- the ragdoll then was the 3D mesh humanoid, and the response of the fabric sensors as the PM.

\subsection{Multimodal pose estimation}
Due to extrinsic properties (\eg strong lighting and occlusion), along with privacy concerns (\ie often RGB is invasive), multimodal solutions to pose estimation have been of recent interest~\cite{hong2015multimodal,sapp2013modec,mehta2017monocular}. Yang~\etal proposed using additional modalities as weak labels for the RGB signal to control latent space representing hand poses with embeddings of heat maps, depth maps, and point clouds~\cite{yang2019aligning}. Zhou~\etal proposed a framework capable of inferring 3D pose information from a 2D image by using an adversarial scheme to train with depth information that was only available for some of the training~\cite{zhou2018adversarial}.
\section{Methodology}\label{sec:methodology}

We proposed a multimodal approach to predict the 3D Mesh of a human laying in bed regardless of the occlusion of blankets. Specifically, the two most informative modalities (i.e. depth and IR) are first fused to generate good initial pose and shape estimation. Then the other two modalities (i.e. PM and RGB) are further fused one by one to refine the result. Let us first describe the human body model (Section~\ref{subsec:SMPL}) and present the proposed framework for single fusion level ($k=0$) (Section~\ref{subsec:single:fusion}), then we will introduce the pyramid fusion scheme as a whole in Section~\ref{subsec:multi:fusion}.

\subsection{The human body model}\label{subsec:SMPL}

We employed the \ac{smpl}~\cite{loper2015smpl} model to encode 3D meshes of humans via two parameters, parameter vectors for pose $\theta$ and shape $\beta$.
Specifically, $\theta$ is 24$\times$3 of scalars for relative rotations of 3D joints in a vector arbitrary to its axes; $\beta$ contains 10 scalars for the magnitudes of a characteristic of the human body like shorter or taller and heavier or lighter. Note that $\beta$ values are PCA shape coefficients for the human 3D mesh shape. Then, the function $\mathcal{M}(\theta, \beta)$ maps the parameters to a 3D body mesh $M\in\mathbb{R}^{N\times3}$, where the number of vertices $N=6,890$. A camera model is also employed to solve the parameters of global rotation $R\in \mathbb{R}^{3 \times 3}$, translation $t\in \mathbb{R}^{2}$, and scale $s\in \mathbb{R}$. Thus, the 3D human body can be represented with a set of mesh parameters ($\theta, \beta$) and camera parameters $\mathcal{C}=(R,t,s)$. Let $x\in \mathbb{R}^{3 \times N}$ be the vertices or joints of the human body mesh $\mathcal{M}(\theta,\beta)$, the projection of $x$ can be expressed as:
$$
\hat{x} = proj(x, \mathcal{C}) = s\Pi(R\mathcal{M}(\theta,\beta)),
$$
where $\Pi$ is the orthographic projection.

\subsection{Single fusion level}\label{subsec:single:fusion}
Given two or more modalities, the goal of each fusion level is to generate accurate pose and shape estimation by leveraging the knowledge captured by the multimodal sensors. However, even with multimodal data, the task of detecting human bodies at rest is still very challenging due to the extreme occlusion of bodies.
To reduce the negative effects of the occlusion from the blanket, we propose a coarse-to-fine estimation process for each fusion level $k$. Specifically, a coarse 3D mesh estimation is first generated from several convolutional and fully-connected layers. Then an attention-based reconstruction module is employed on the coarse mesh to generate the uncovered modalities. Finally, the uncovered modalities are fused to update the current estimation in a cyclic fashion.

Taking the first fusion level ($k=0$) as an example, Figure~\ref{fig:proposedframework} depicts the end-to-end framework to infer the 3D human body and camera from depth and IR data. Specifically, depth and IR images \textit{with cover} ( ${I}_{\text{D}}\oplus{I}_{\text{IR}}$) are first concatenated and fed to a shared regression model $g$ to generate a coarse prediction of SMPL parameters $\Theta_0 =\{ \theta_0, \beta_0\}$ and camera parameters $\mathcal{C}_0$. Therefore, we can generate the mesh corresponding to the regressed parameters, $M_{0}=\mathcal{M}(\theta_0, \beta_0)$. The regression model we used has the same design as SPIN~\cite{kolotouros2019learning}, and is supervised by minimizing the combination of a reprojection loss $\mathcal{L}_{2D}^0$ on the 2D joints, a 3D joints loss $\mathcal{L}_{3D}^0$, and a smpl loss $\mathcal{L}_{smpl}^0$:
\begin{equation}
    \mathcal{L}_{2D}^0=||\hat{J}_0 - J_{gt}||_2,
\end{equation}
where $\hat{J}_0 = proj(J_0, \mathcal{C})$ are the projection of the 2D joints $J_0$, and $J_{gt}$ are the ground truth 2D joints. $||\cdot||_2$ denotes L2-norm. Superscripts of $\mathcal{L}$ denote losses for coarse (\ie 0) or fine (\ie 1) stage.
\begin{equation}
    \mathcal{L}_{3D}^0=||\hat{X}_{0} - X_{gt}||_2,
\end{equation}
where $\hat{X}_0$ are the 3D joints of $M_0$, and $X_{gt}$ are the ground truth 3D joints.
Since we do not have the ground truth of SMPL parameters $[\theta,\beta]$, we follow SPIN to generate an optical mesh $M_{opt} = \mathcal{M}(\theta_{opt},\beta_{opt})$ on top of regressed parameters using an optimization-based approach~\cite{bogo2016keep}. Given $\Theta_{opt} =\{ \theta_{opt}, \beta_{opt}\}$, then the smpl loss can be expressed as
\begin{equation}
    \mathcal{L}_{smpl}^0=||M_{0}- M_{opt}||_1 + ||\Theta_{0}- \Theta_{opt}||_2,
\end{equation}
where $||\cdot||_1$ is the L1-norm.

We then reconstruct the \textit{uncovered} depth and IR images ( ${\hat{I}}^{un}_{\text{D}},{\hat{I}}^{un}_{\text{IR}}$) based on the inferred parameters. 
To learn a precise mapping from \textit{cover} to \textit{uncover}, we employ an attention-based reconstruction module to reduce the noise caused by coverage, and hence reinforce the model to focus on the body part. The details of reconstructioin will be introduced in Section~\ref{subsec:reconstruction}. Finally, the reconstructed depth and IR images are sent back to the shared regression model $g$ and output the fine estimation of human body and camera parameters ($\theta_1, \beta_1, \mathcal{C}_1$). Similarly, we can get $\mathcal{L}_{2D}^1$, $\mathcal{L}_{3D}^1$, and $\mathcal{L}_{smpl}^1$.
Therefore, the overall objective function for shared regressor $g$ can be expressed as:
\begin{equation}
    \mathcal{L}_{g}= \sum_{i=\{0,1\}} \bigg( \mathcal{L}_{2D}(i)+\mathcal{L}_{3D}(i)+\mathcal{L}_{smpl}(i) \bigg).
\end{equation}

\subsection{Attention-based reconstruction}~\label{subsec:reconstruction}
Given the coarse predictions of human body parameter $\Theta_0 =\{ \theta_0, \beta_0\}$ and camera parameter $\mathcal{C}_0$, we reconstruct the \textit{uncovered} depth and IR images ( ${\hat{I}}^{un}_{\text{D}},{\hat{I}}^{un}_{\text{IR}}$) by employing an attention-based reconstruction module. Specifically, the 3D vertices of the human mesh model $\mathcal{M}(\theta_0, \beta_0)$ gets projected to the 2D image plane and made into a boolean mask ${\hat{I}}_m$ representing the body’s silhouette. The mask loss $\mathcal{L}_{mask}$ can be defined as:
\begin{equation}
    \mathcal{L}_{mask} = ||{\hat{I}}_m-{I}_m^{gt}||_1,
\end{equation}
where ${I}_m^{gt}$ is the ground truth mask.

Then ${\hat{I}}_m$ is used to segment out the human body in [${I}_{\text{D}}, {I}_{\text{IR}}$] and forwarded into a decoder $d$ along with features of $g$ to generate the uncovered IR and depth images:
\begin{equation}
    ({\hat{I}}_{\text{D}}^{un},{\hat{I}}_{\text{IR}}^{un})= d~({\hat{I}}_m \odot {I}_{\text{D}}, {\hat{I}}_m \odot {I}_{\text{IR}}, ~g({I}_{\text{D}}\oplus{I}_{\text{IR}})),
\end{equation}
where $\odot$ denotes element-wise product. The decoder $d$ consists of a small mount of convolutional and upsampling (\ie pixel shuffle~\cite{shi2016real}) layers. The structure details of $d$ are described in Section~\ref{subsec:implementation}.

To further reinforce the model to pay attention to body part, we add reconstruction loss (\ie L1 loss) on the masked depth and IR image.
Given the ground truth of uncovered depth and IR images ($I_\text{D}^{un},I_\text{IR}^{un}$), the objective function of $d$ can be expresses as:
\begin{equation}
\begin{split}
    \mathcal{L}_{d} &= \mathcal{L}_{mask} + \mathcal{L}_{recon}\\
    &= \mathcal{L}_{mask} + \sum_{t = \{\text{D}, \text{IR}\}} ||\hat{I}_t^{un}- I_t^{un}||_1 \odot I_m^{gt}.
\end{split}
\end{equation}

Note that we do not recover RGB in any fusion levels, since bodies are usually under extreme occlusion and almost no clues are provided to recover RGB data.

\subsection{Multimodal fusion}\label{subsec:multi:fusion}
Given inputs aligned in view but a difference in the type of modality, the overarching goal is to efficiently model the correlations across modalities. First, we evaluated each modality independently to support the hypothesis that performances will vary-- the question then is how to make each modality best complement each other.
All the while, recalling the inherent challenges posed by high variations in lighting and from the vast occlusion from blankets (\ie covers), it was to no surprise that RGB performs the poorest on covered poses, while PM has almost the same performance across different cover types. However, PM is vulnerable to missing information, which tends to cause inaccurate predictions. Comparing to RGB and PM, depth and IR enable better human \emph{in-bed} pose estimation despite of occlusions.



To effectively fuse the different modalities in a way that best leverages the knowledge captured by the multimodal sensors, we propose a pyramid scheme to fuse the different modalities at different levels (Fig.~\ref{fig:pyramid}). When training each fusion level, the learned model weights of previous level are freezed to enable the fine-tuning of the estimation. The first level (\ie $k=0$) assumes the mean parameters for the 3D mesh (\ie $(\Vec{\beta}^0, \Vec{\theta}^0)$), with each proceeding level passed the predicted parameters from the level prior. Hence, the process starts at the bottom of the pyramid, passing in ${I}_{\text{D}}$ and ${I}_{\text{IR}}$: the top performing modalities (Table~\ref{tbl:result_ablation_single_mod}), and also the modalities that provide shape and other local information in the dark and while under the covers (\ie occluded). 
At the next level (\ie $k=1$), the parameters of the 3D mesh $(\Vec{\beta}^1, \Vec{\theta}^1)$ are inferred by the learned mappings of the previous layer $g^{k-1}(\cdot)$ (Fig.~\ref{fig:proposedframework}), from the input ${I}_{\text{D}}\oplus{I}_{\text{IR}}$ (\ie concatenate tensors across the channel dimension). Hence, the process at the first level can be expressed as $g^0\xrightarrow{}(\Vec{\beta}^1, \Vec{\theta}^1)$. 
The input of $k=1$ is the three element tuple $({I}_{\text{D}}\oplus{I}_{\text{IR}}\oplus{I}_{\text{PM}}, \Vec{\beta}^1, \Vec{\theta}^1)$; the input to learn the mapping $g^1\xrightarrow{}(\Vec{\beta}^2, \Vec{\theta}^2)$ which, again, go to the next level. Specifically, $l=2$ is the next, final level that is fed all four modalities to output the final 3D mesh prediction. The pseudocode of the training process are listed in Algorithm~\ref{alg:pyramid_fusion}.

% ...


  


\begin{algorithm}[t]
\SetKwInput{KwInput}{Input}                
\SetKwInput{KwOutput}{Output}              
\DontPrintSemicolon
  \KwInput{Images, ${I}=\{{I}_{\text{RGB}}, {I}_{\text{IR}}, {I}_{\text{D}}, {I}_{\text{PM}}\}$; average shape and pose parameters (\ie $(\Vec{\beta}^0, \Vec{\theta}^0)$), which was found via train set.}
  \KwOutput{Prediction shape and pose $(\Vec{\beta}, \Vec{\theta})$ having been refined at each of $k$ levels of fusion.}
  \BlankLine
  ${I}_{in}={I}_{\text{D}} \oplus {I}_{\text{IR}}$ 
  \tcc{Until change in error falls below $t$ or for $B$ batches.}
    \While{$t\leq T\textbf{ and } b\leq B$ }{

        \tcc{2 MODALITIES}
        
        ${I}_{0}={I}_{in}$
        
        $\Vec{\beta}^1, \Vec{\theta}^1={g}^0({I}_{0}, \Vec{\beta}^0, \Vec{\theta}^0)$

    \tcc{Fuse in PM image using the predicted $\beta$ and $\theta$ from level-0.}
    \If{$K>0$}
    {   
        \tcc{3 MODALITIES}
        freeze parameters in $g^0$
        
    	${I}_{1}={I}_{in} \oplus {I}_{\text{PM}}$

        $\Vec{\beta}^2, \Vec{\theta}^2={g}^1({I}_{1}, \Vec{\beta}^1, \Vec{\theta}^1)$
    }
    \tcc{Fuse in RGB image using the predicted $\beta$ and $\theta$ from previous level.}
    \If{$K>1$}
    {       \tcc{4 MODALITIES}
            freeze parameters in $g^0$, $g^1$
            
            ${I}_{2}={I}_{in} \oplus {I}_{\text{PM}} \oplus {I}_{\text{RGB}}$
             
            $\Vec{\beta}^3, \Vec{\theta}^3={g}^2({I}_{2}, \Vec{\beta}^2, \Vec{\theta}^2)$
    }
    
    $\Vec{\beta}, \Vec{\theta} = \Vec{\beta}^{K+1}, \Vec{\theta}^{K+1}$
  }
 \caption{\textbf{Algorithm of the proposed.} Multimodal process analogous to the transcending of the pitched-side of a pyramid.}
 \label{alg:pyramid_fusion}
\end{algorithm}

\begin{figure*}
\begin{center}
\includegraphics[width=\textwidth]{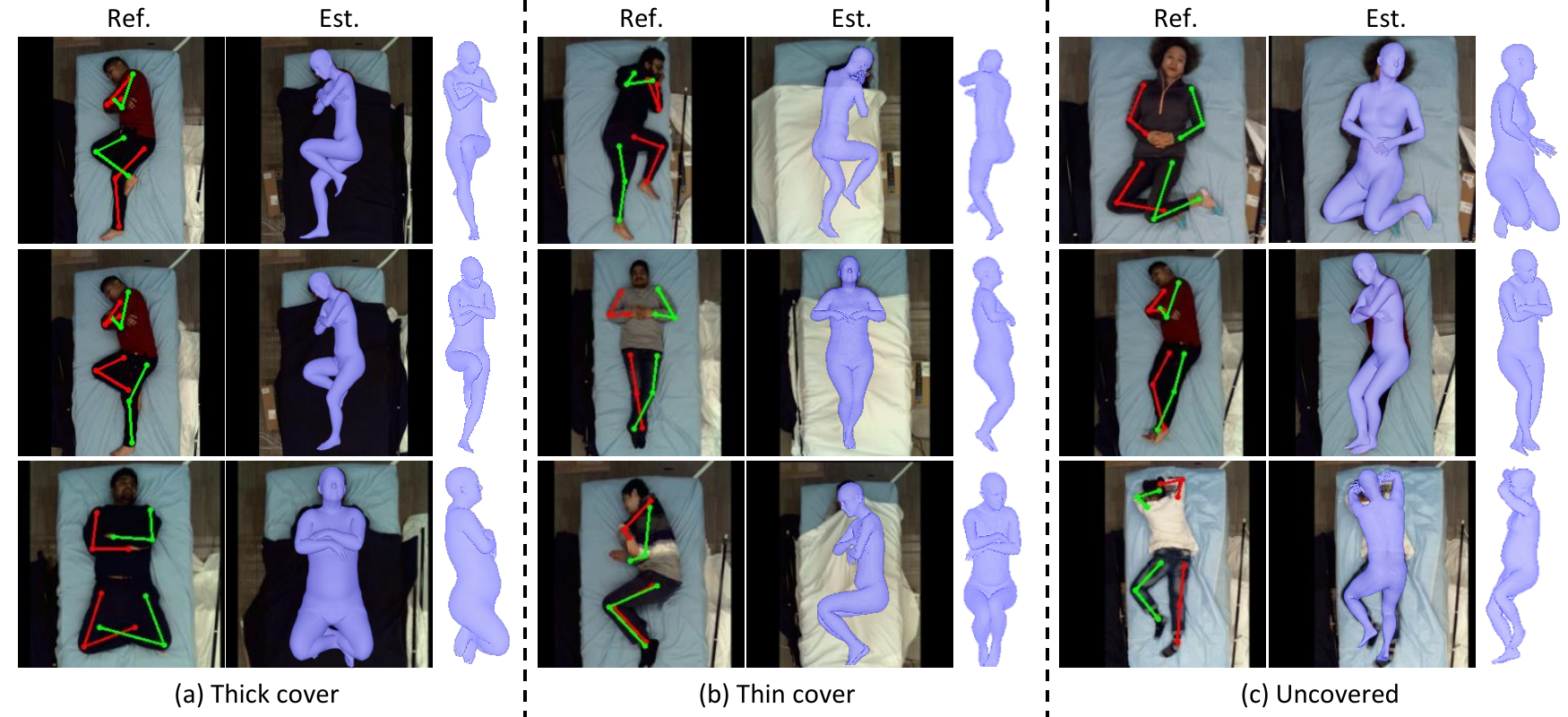}
\end{center}
  \caption{\textbf{Qualitative results.} Samples randomly selected to represent different cover types (\ie thick, thin, and no cover). In reference image, left arms and legs are labeled in {\color{green}green}, right body parts are labeled in {\color{red}red}.}
\label{fig:qualitative_results}
\end{figure*}

\subsection{Implementation}\label{subsec:implementation}
\textbf{Dataset:} All the modalities are well aligned in view. The input images are cropped and resized to $224\times224$. The training images were augmented using random scaling, rotation, and horizontally flipping. Since the joint definition of SMPL does not alighed with joints in SLP~\cite{liu20120simultaneously}, we follow \cite{kanazawa2018end,bogo2016keep} to use a regressor to obtain the 14 joints from the reconstructed mesh.

\textbf{Architecture:} Our regression network $g$ has the same structure as \cite{kanazawa2018end,kolotouros2019learning}, which consists of a ResNet-50 network~\cite{he2016identity} to encoder images features and a 3D regression module to predict human model and camera parameters of 85 dimensions. The reconstruction module $d$ consists of 2 branches. The first branch extracted features from the masked modality using one residual block~\cite{he2016identity} with stride 2. 
The second branch takes image features (\ie the output of encoder in $g$) as input, and pass them through four convolutional layers, each followed by a pixel shuffle layer. The output of both branch has dimention of ${(112,112,64)}$. Finally, the output of these two branches are concatenated to feed in a residual block, followed by a pixel shuffle and a convolutional layer to output the final recovered modality $I \in \mathcal{R}^{224\times224\times1}$. 
Optimization was done with ADAM with a learning rate of $1 \times 10^{-5}$. The model was trained
with a batch size of 24 and for a total epoch of 100 epochs. Implementation was done using PyTorch. Training took about 3 days with a Nvidia TITAN-XP GPU.

\section{Experiments}\label{sec:experiments}

\subsection{Data}
As shown in Fig.~\ref{fig:pyramid}, the data used consists of multimodal paired data of subjects posing in bed, both without and under the covers~\cite{liu20120simultaneously, liu2019seeing}.\footnote{\url{https://web.northeastern.edu/ostadabbas/2019/06/27/multimodal-in-bed-pose-estimation/}.} This data collection was dubbed the \acf{slp} dataset. \ac{slp} was collected in two indoor settings, with seven participants in a hospital setting and 102 participants at home setting. We used the first 85 subjects of the home setting for training and evaluated on the remain 22 subjects.
For each instance, four imaging modalities were captures. Specifically, RGB (\ie regular webcam), LWIR (\ie FLIR LWIR camera), depth (\ie Kinect v2), and pressure map (\ie Tekscan pressure sensing map); syntactically, the set of multimodal images ${I}=\{{I}_{\text{RGB}}, {I}_{\text{IR}}, {I}_{\text{D}}, {I}_{\text{PM}}\}$. Each modality shares the same field of view as the others in each instance (\ie the data are aligned). Also, instances are represented by three conditions, which again are aligned: no cover, covered with a bed sheet (\ie \emph{cover 1}), and covered with a blanket (\ie \emph{cover 2}).



\subsection{Comparison with state-of-the-art methods}
We report results on 3D pose estimation and compare with state-of-the-art methods including both types of methods designed for general human pose estimation (\ie HMR~\cite{kanazawa2018end}, SPIN~\cite{kolotouros2019learning}), and \emph{in-bed} pose estimation (\ie Bodies At Rest~\cite{clever2020bodies}).

\begin{table*}[t!]
\centering
\caption{\textbf{Quantitative benchmark,} MPJPE (right) and reconstruction error (left). Lower scores are better.}
\label{tbl:result_joint}
 \begin{tabular}{r|cccccccc}
 \toprule
    &\multicolumn{3}{c}{\textbf{MPJPE}}&
    & \multicolumn{3}{c}{\textbf{Rec. Error}} \\
    \cline{2-4}\cline{6-8} 
    & Cover 2 & Cover 1 & Uncover && Cover 2  & Cover 1 & Uncover\\
    \midrule
    HMR~\cite{kanazawa2018end} & 109.83 & 106.64 & 100.86 && 100.03 & 98.42& 93.04\\
    HMR (4 modalities) & 94.16 & 93.32 & 93.05 && 87.12 & 86.54 & 86.76 \\
    \ac{spin}~\cite{kolotouros2019learning} & 99.36 & 98.37 & 87.77 && 90.00 & 88.98 & 78.70\\
    \ac{spin} (4 modalities) & 86.40 & 85.72 & 83.59 && 78.20 & 77.57 & 74.66 \\
    Bodies At Rest~\cite{clever2020bodies} & 96.10 & 96.84 & 96.10 && 84.37 & 84.12 & 84.19\\
    Bodies At Rest (4 modalities) & 86.98 & 86.75 & 86.83 && 78.78 & 78.51 & 78.40 \\
    \midrule
    Ours ($k$ = 0) & 82.69 & 82.08 & 80.39 && 74.41 & 73.56 & 71.83 \\
    Ours ($k$ = 1) & 81.65 & 81.24 & 80.44 && 72.49 & 71.95 & 71.39\\
    Ours ($k$ = 2) & \textbf{80.21} & \textbf{79.92} & \textbf{78.80} && \textbf{71.80} & \textbf{71.44} & \textbf{70.65}\\
\bottomrule
\end{tabular}
\end{table*}

\begin{table}[t!]
\centering
\caption{\textbf{Foreground-background segmentation.} Higher scores are better. \# of papamrters of models are also listed.}
\label{tbl:result_FB_segmentation}
\begin{adjustbox}{max width=\linewidth}
\begin{tabular}{c|ccc}
    & Accuracy & F1 & \# of Parameters\\
    \toprule
    HMR~\cite{kanazawa2018end}  & 92.11 & 0.849 & \textbf{25.72} \\
    \ac{spin}~\cite{kolotouros2019learning} & 92.78 & 0.853 & \textbf{25.72}\\
    Bodies At Rest~\cite{clever2020bodies} & 92.10 & 0.840 & 140.60\\
    Ours  & \textbf{93.54} &\textbf{0.867}& 39.94 \\
\end{tabular}
\end{adjustbox}
\end{table}

\textbf{3D joint location estimation.} \label{sec:joint_estimation}
We first show results on 3D joint estimation using two common evaluation metrics: the mean per joint position error (\textit{MPJPE}) and \textit{Reconstruction} error. MPJPE measures the mean Euclidean distance between ground truth and predicted joints. The Reconstruction error is MPJPE after rigid alignment of the prediction with ground truth using Procrusters Aynalysis~\cite{gower1975generalized}.
To evaluate the proposed, we use \emph{uncover} (\ie no covers), \emph{Cover 1} (\ie thin covers), and \emph{Cover 2} (\ie thick covers) data of SLP, reporting performances separately. Notice that the more the coverage the greater the error, as would be expected. 

We report results of the proposed method in different fusion levels (\ie $k$ = 0, 1, 2). Considering other methods only use either RGB or PM to estimate human pose and shape, we first report results of their method using single modality according to their paper. Besides, we also show fusion results of their method by concatenating all four modalities (see Table~\ref{tbl:result_joint}). Specifically, two general human pose estimation methods~\cite{kanazawa2018end,kolotouros2019learning} are evaluated using single RGB modality and the fusion of all four modalities separately. The \emph{in-bed} pose estimation methods~\cite{clever2020bodies} are evaluated in both PM and fusion settings. As shown in Table~\ref{tbl:result_joint}, the fusion of four modalities largely boost the performance of each, and balanced the estimation error among three cover types (\ie uncover, cover 1, and cover2).
Still, the proposed method out-performs state-of-the-art works by a large margin. We achieve better performance even in the first fusion level, where only IR and depth images are fused. Then in the next level ($k$ = 1), the occlusion-invariant information from PM further pushes the performance by correcting the estimation of the covered part. We can see from Table~\ref{tbl:result_joint} that the error of level $k$ = 1 dropped for cover 1 and cover 2 settings, but slightly increased in the uncovered setting. Finally, at the final level ($k$ = 2), the RGB is fused to provide additional details and an accurate shape for the uncovered part. Hence, we can see more performance improvement in the uncover setting instead of the other two.

\begin{figure}
\begin{center}
\includegraphics[width=\linewidth]{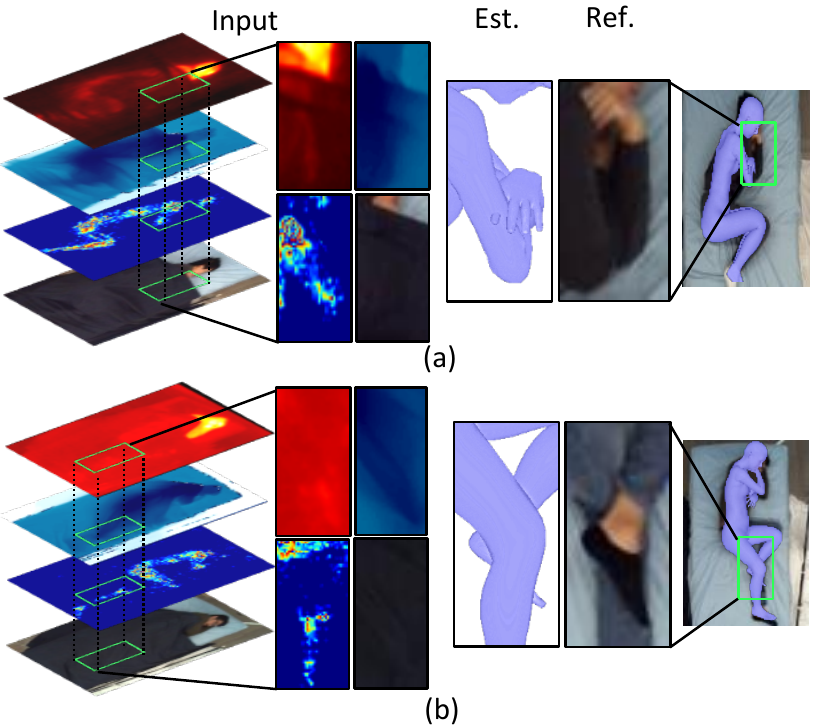}
\end{center}
  \caption{\textbf{Failure cases.} Notice the pose ambiguities for cases that lacked information across all four modalities. Levels $K=0$, $K=1$, and $K=2$) (\ie IR, PM, and RGB respective of (a) fail to provide useful information for cuddled limbs under thick coverage, while PM provides wrong clues, hence leading to the false arm estimation. (b) False leg pose due to thick coverage and crossed legs.}
\label{fig:failure_cases}
\end{figure}

We then show qualitative results of our method under different covering settings in Fig.~\ref{fig:qualitative_results}. The results show that our method performs well on challenging cases (\ie thick cover). Besides, with the help of multi-modalites, our method can also make correct estimation on confusing poses (\ie poses are hard to recognize even in reference image (see top two rows of Fig.~\ref{fig:qualitative_results} (a)). A large variety of results can be found in the Supplementary Material.
Failure cases are shown in Fig.~\ref{fig:failure_cases}. Pose ambiguities are mostly caused by lack of information in all modalities.

\textbf{Human body segmentation.}
Since there is no ground truth mesh for the dataset, it is hard to evaluate the quality of estimated shapes. Instead, we use the silhouette segmentation to implicitly evaluate 3D shapes. 
Results are reported with segmentation accuracy and F1 score of projected mesh. 

In this section, we use only IR and depth modalities to demonstrate the superior of our model even without pyramid fusion scheme and further prove the effectiveness of proposed attention-based reconstruction module. Qualitative results for foreground-background segmentation are shown in Table~\ref{tbl:result_FB_segmentation}. We also compare the number of parameters used in different models in Table~\ref{tbl:result_FB_segmentation}. We have larger model than HMR and SPIN because of the extra attention-based reconstruction model. However, comparing to Bodies At Rest, who also includes a similar recovering module, our model has much smaller size.

\begin{table}[t!]
\centering
\caption{\textbf{Ablation study.} Comparison of pyramid fusion (\textit{P}) and the contribution of each modality (evaluated by Reconstruction errors in mm).}
\label{tbl:result_ablation_single_mod}
\begin{adjustbox}{max width=0.9\linewidth}
\begin{tabular}{c|cccc}
    \toprule
    & Cover2 & Cover1 & Uncover \\
    \midrule
    RGB  & 90.00 & 88.98 & 78.70 \\
    PM & 79.47 & 79.40 & 79.38\\
    Depth  & 79.38 & 79.48 & 75.16 \\
    IR & 77.51 & 76.73 & 76.03\\
    \midrule
    IR+Depth (w/o \textit{P}) & 80.48 & 79.10 & 77.72 \\
    IR+Depth+PM (w/o \textit{P}) & 78.74 & 78.35 & 77.78\\
    IR+Depth+PM+RGB (w/o \textit{P}) & 78.02 & 77.60 & 75.12 \\
    IR+Depth+PM+RGB (w/~ \textit{P}) & 76.03 & 75.36 & 74.83 \\
    \bottomrule
\end{tabular}
\end{adjustbox}
\end{table}

\begin{figure}
\begin{center}
\includegraphics[trim=0in 0in 0in 0in,clip,width=\linewidth]{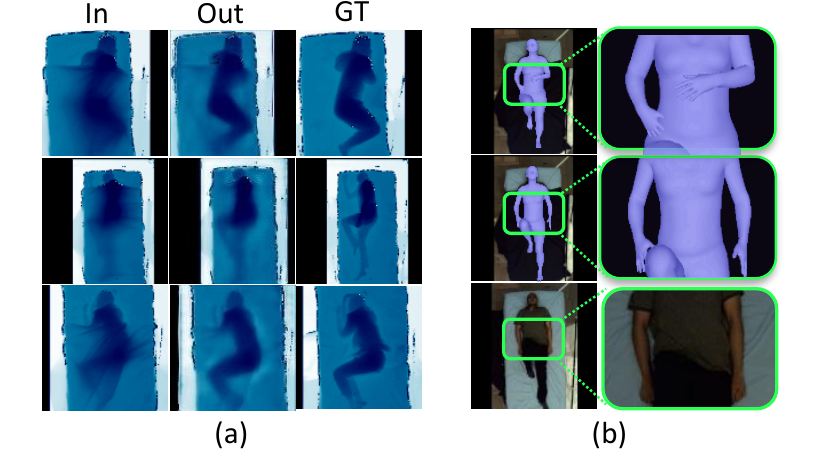}
\end{center}
  \caption{\textbf{Ablation study for attention-based reconstruction module.} Referring to \textbf{(a)}, notice the depth data of covered input (\emph{left}), reconstructed output (\emph{middle}), uncovered GT (\emph{right}); then, for \textbf{(b)} shows estimated mesh before (\emph{top}), and after (\emph{middle}) attention-based reconstruction module, with the reference shown in the \emph{bottom}.}
\label{fig:ablation_uncovering}
\end{figure}

\subsection{Ablation study}
The components of our approach are evaluated in this section. We first show the effectiveness of our pyramid fusion scheme and analyse the contribution of each modality. Then we demonstrate the superior of the proposed attention-based reconstruction module.

\textbf{Effect of pyramid fusion scheme.}
To gain insight of pyramid fusion scheme and the contribution of each modality, we conduct experiments using 1) single modality only, 2) fusion with pyramid, and 3) fusion without pyramid scheme, separately (see Table~\ref{tbl:result_ablation_single_mod}).
Notice that RGB data is difficult to recover when all the body parts are covered. For fair comparison, we remove the reconstruction modules for all the experiments in this section. 
Results shown that RGB data performs the poorest on covered poses as expected. The performance of PM data are invariant to cover types. However, we notice that PM data does not outperform RGB data for uncovered poses, which demonstrates that PM data are inaccurate and vulnerable to missing information. In general, depth and IR data are more informative and can recover better knowledge for the \emph{in-bed} pose estimation task than the other two modalities, though depth data may sometimes be affected by occlusions.

Unsurprisingly, by fusion modalities, we can further improve the performance of the model, since it gradually get access to more information. Finally, we show that fusion via pyramid scheme performs the best, and thus validates its contribution to state-of-the-art performance.

\textbf{Effect of attention-based reconstruction module.}
We first show some samples of the reconstructed image in Fig.~\ref{fig:ablation_uncovering} (a) to validate the effectiveness of the attention-based reconstruction module. Note that the two modalities, that are affected by coverage the most, are RGB and depth. Since RGB can not be recovered, we compared the reconstructed depth images (middle) with covered input (left), and uncovered ground truth (right) in Fig.~\ref{fig:ablation_uncovering} (a). We can see that the reconstructed depth images can provide more pose information than the original covered images, which leads to the improvement of performance.
Furthermore, we compare the pose and shape estimation before and after reconstruction in Fig.~\ref{fig:ablation_uncovering} (b). It is shown that the prediction of pose and shape are improved after reconstruction. During experiments, We also observed an average of 2mm performance boost with proposed reconstruction module.


\section{Conclusion}
We proposed a multimodal framework to predict 3D meshes of humans in bed regardless whether it is covered or not. To the best of our knowledge, we are the first to infer \emph{in-bed} poses and shapes from multiple modalities. After evaluating each modality independently, a pyramid scheme is proposed to effectively fuse different modalities in a way that best complement each other. The two most informative modalities (\ie depth and infrared) are first fused to generate good initial estimations. Then pressure map and RGB are further fused to provide occlusion-invariant knowledge for covered body part, and accurate shapes for the uncovered part, respectively. Furthermore, an attention-based reconstruction module is proposed to further reduce the negative effect of blanket coverage. We explore the contribution of each modality and show the effectiveness of our pyramid fusion scheme and attention-based reconstruction module.

{\small
\bibliographystyle{ieee_fullname}
\bibliography{egbib}
}

\end{document}